\documentclass[final,5p,times,twocolumn]{elsarticle}

\usepackage{amssymb}
\usepackage{amsmath}

\usepackage{multirow}
\usepackage{url}
\usepackage{subfigure}
\usepackage{textcomp}
\usepackage{algorithm}
\usepackage{algpseudocode}
\usepackage{flushend}

\journal{}

\begin{document}

\begin{frontmatter}



\title{An ARGoS plug-in for the Crazyflie drone}


\author[myaddress1]{Daniel~H.~Stolfi\corref{mycorrespondingauthor}}
\cortext[mycorrespondingauthor]{Corresponding author}
\ead{daniel.stolfi@uni.lu}

\author[myaddress1,myaddress2]{Gr\'egoire~Danoy}
\ead{gregoire.danoy@uni.lu}


\address[myaddress1]{Interdisciplinary Centre for Security, Reliability and Trust (SnT), University of Luxembourg}
\address[myaddress2]{FSTM/DCS, University of Luxembourg}



\begin{abstract}

We present a new plug-in for the ARGoS swarm robotic simulator to implement the Crazyflie drone, including its controllers, sensors, and some expansion decks.
We have based our development on the former Spiri drone, upgrading the position controller, adding a new speed controller, LED ring, onboard camera, and battery discharge model.
We have compared this new plug-in in terms of accuracy and efficiency with data obtained from real Crazyflie drones.
All our experiments showed that the proposed plug-in worked well, presenting high levels of accuracy.
We believe that this is an important contribution to robot simulations which will extend ARGoS capabilities through the use of our proposed, open-source plug-in.

\end{abstract}



\begin{keyword}
Crazyflie \sep ARGoS \sep computer simulations \sep sensors \sep swarm robotic 



\end{keyword}

\end{frontmatter}


\section{Introduction}
\label{sec:intro}

Conducting real world swarm intelligence experiments could be challenging, especially during the early stages of the research approach.
Uncertainties related to the tested algorithms, positioning system accuracy, discrepancies between the mathematical model and actual robots, might end up in catastrophic collisions, which are even worse when using drones~\cite{Harker2022}.
Economic consequences are not the only to be taken into account as people's safety is also a concern~\cite{Zhu2021a}.

Computer simulations~\cite{Symeonidis2022} have been used as a tool not only for experimenting with robotic systems without taking any risk, but also for performing thousands to millions of tests in a very short time compared to real world experiments.
ARGoS~\cite{Pinciroli:SI2012} is an open-source swarm robotic simulator featuring a modular multithread architecture.
It is capable of efficiently simulating multi-robot swarms, including sensors, actuators, and communications.
Thanks to ARGoS' modular architecture, it is possible to extend its features creating new robots, sensors, actuators, physic models, etc.
ARGoS provides an optional 3D graphical environment to visualise the simulation, and its performance has been compared to other simulators in~\cite{Pitonakova2018}.

\begin{figure}[t]
   \centering
   \includegraphics[width=0.5\linewidth]{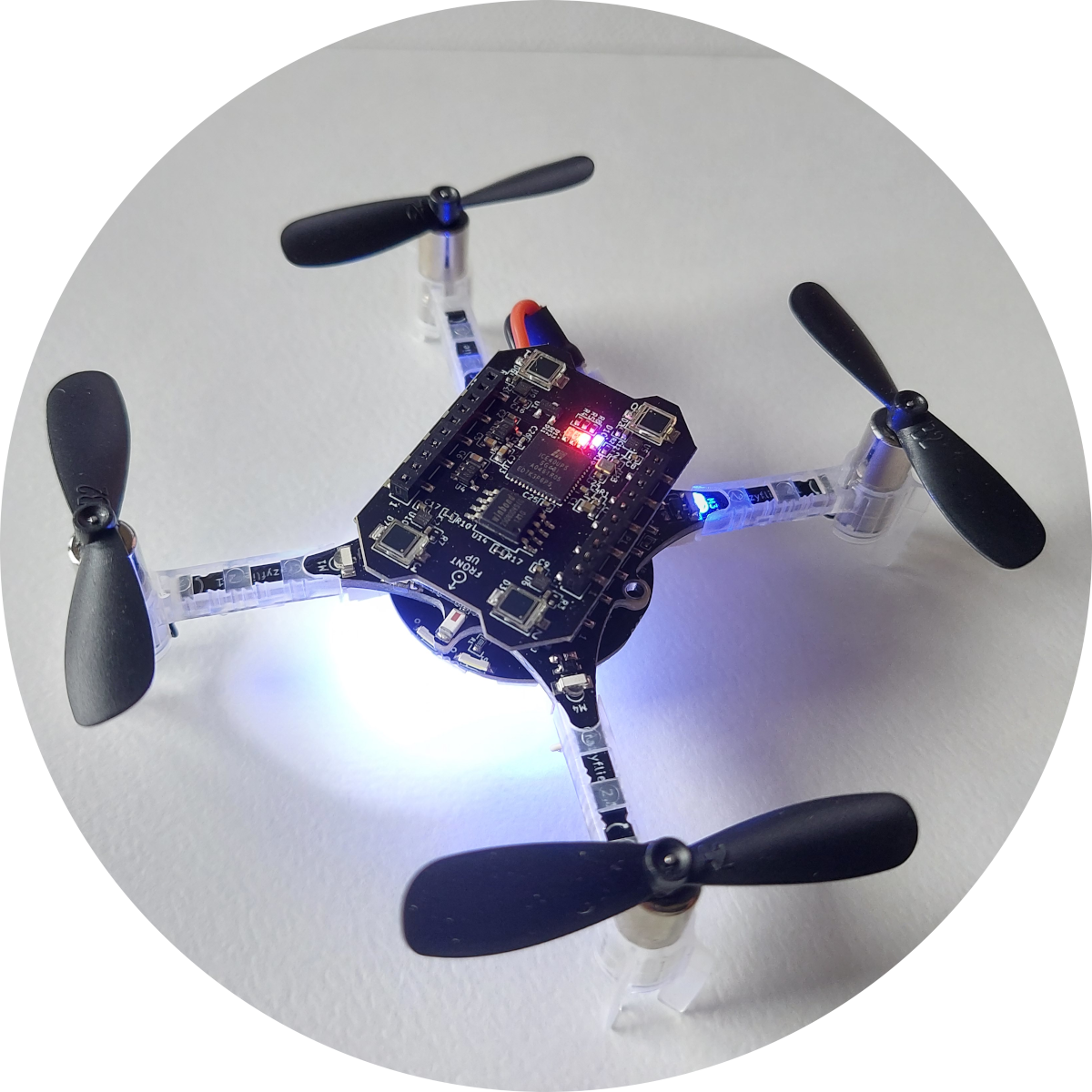}
   \caption{The Crazyflie drone.}
   \label{fig:crazyflie}
\end{figure}

The current ARGoS version (3.0.0-beta59) includes the following robot plug-ins: E-Puck, Eye-Bot, Foot-Bot, Pi-puck, Spiri, among others.
We have conducted some research works using the Spiri model for 3D drone formations~\cite{Stolfi2023a,Stolfi2023b}.
However, we have found that the dynamic behaviour of Spiri does not match our experimental hardware: Crazyflie drones (Figure~\ref{fig:crazyflie}).
The need to model our drones in ARGoS became evident when we decided to validate our simulations using our real drones.
Hence, we have developed the Crazyflie drone plug-in for the ARGoS simulator, accurately modelling its onboard sensors and actuators plus some of its expansion decks.
In addition we have modified the PD (proportional-derivative) controllers (position and velocity) to approximate the simulated drone trajectories to those observed in the real drones.
The plug-in source code is freely available at \url{https://gitlab.uni.lu/adars/crazyflie} to be used by researchers in their experiments involving Crazyflie drones and ARGoS.

The rest of this paper is organized as follows.
In the next section, we review the state of the art related to our work. 
In Section~\ref{sec:proposal} the proposed Crazyflie plug-in and its sensors and actuators are presented.
Our experiments and results are discussed in Section~\ref{sec:results}.
And finally, Section~\ref{sec:conclusion}, brings conclusions and future work.

\section{Related Work}
\label{sec:relatedwork}

In this section we review some related works comprising different plug-ins for the ARGoS simulator as well as some of the quadcopter models available in other simulators.

Different robot plug-ins can be found in the ARGoS ecosystem.
They have been focused mainly to wheeled robots such as the E-Puck robot~\cite{Floreano2010}, the E-Puck2 robot~\cite{Stolfi2023}, the Thymio~\cite{Mondada2017} robotic platform, and the Khepera IV~\cite{Pinciroli2016}.
Other robots models have been also implemented in ARGoS, for example the Kilobot~\cite{Pinciroli2018} which uses vibrational motors and the BuilderBot robot~\cite{Zheng2021} which is made of Stigmergic Blocks.
All these develops as well as ours, use the ARGoS plug-in architecture that allows extending its functionality without compiling the simulator's source code.

Quadcopters have been modelled in diverse simulation platforms.
These devices are particularly vulnerable to accidental collisions during the early stages of a project and its test phases.
In ARGoS we can find a plug-in of the Spiri drone from Spiri Robotics~\cite{SpiriRobotics2024}; in the Webots~\cite{Webots2024} simulator, the DJI Mavic 2 PRO and also the Crazyflie drone; and in the CoppeliaSim (formerly V-REP)~\cite{CoppeliaRobotics2024}, a generic quadcopter is available which can also be modified to implement other different models.

We propose a new plug-in for the ARGoS simulator to allow researchers to experiment with Crazyflie drones, not only planning and executing different trajectories but also simulating visual communications and via radio link.
Additionally, a realistic battery discharge model is provided that matches the real drone flying autonomy as well as an optional onboard camera.
In the following sections we describe this novel plug-in, the implemented sensors, actuators and controllers, and also validate them via experiments with actual drones.

\section{Crazyflie plug-in for ARGoS}
\label{sec:proposal}

The development of the Crazyflie drone began in 2009 by a group of designers who later founded the Bitcraze AB~\cite{Bitcraze2024} company.
The Crazyflie drone is a small open-source modular quadcopter which uses its printed circuit board as the mechanical frame and has its four motors physically and electronically plugged.
It supports different extension decks allowing three kinds of positioning systems, optical navigation, camera, LED ring, etc.

The Crazyflie drone weights 27g and its dimensions are 92x92x29 mm without propellers.
It is equipped with an STM32F405 microcontroller, an nRF51822 for radio and power management, a micro-USB connector, on-board LiPo charger, and 8KB EEPROM.
It also features a 3-axis accelerometer/gyroscope (BMI088) as well as a pressure sensor (BMP388).
The specified flight time with stock battery is 7 minutes.

Our implemented Crazyflie plug-in is based on the Spiri robot plug-in developed by Carlos Pinciroli for the ARGoS simulator~\cite{Pinciroli:SI2012}.
We have implemented a bottom LED, velocity and position controllers, onboard camera, and the battery discharge model, all corresponding to the real drone.
It is also possible to use the ARGoS Range and Bearing sensor/actuator to communicate between drones.
This new plug-in can be used in ARGoS by programming and compiling the controller's C++ code, although Lua is also supported.
In the next sections we describe the features available in our Crazyflie plug-in.

\subsection{Body and LEDs}

\begin{figure*}[!t]
   \centering
   \subfigure{\includegraphics[width=0.22\linewidth]{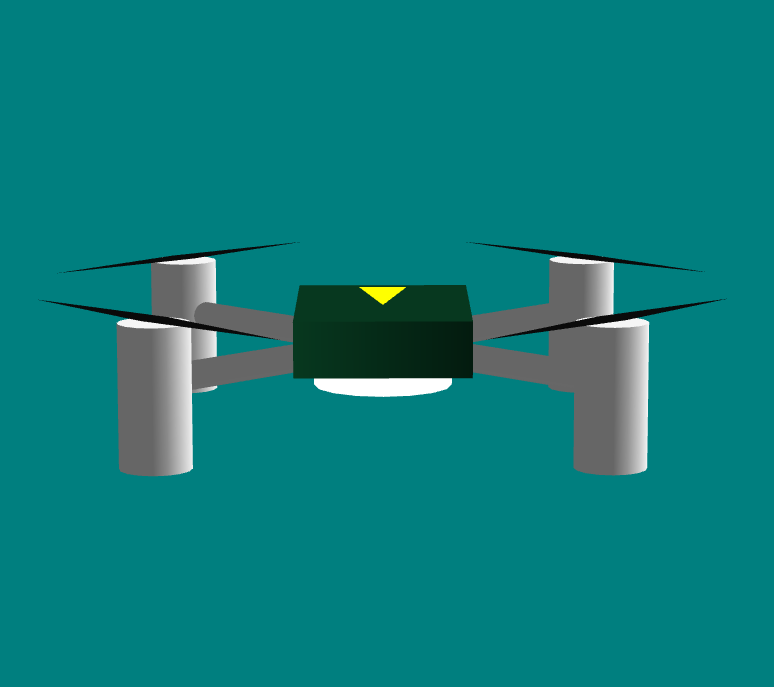}\label{fig:led_white}}
   \hfil
   \subfigure{\includegraphics[width=0.22\linewidth]{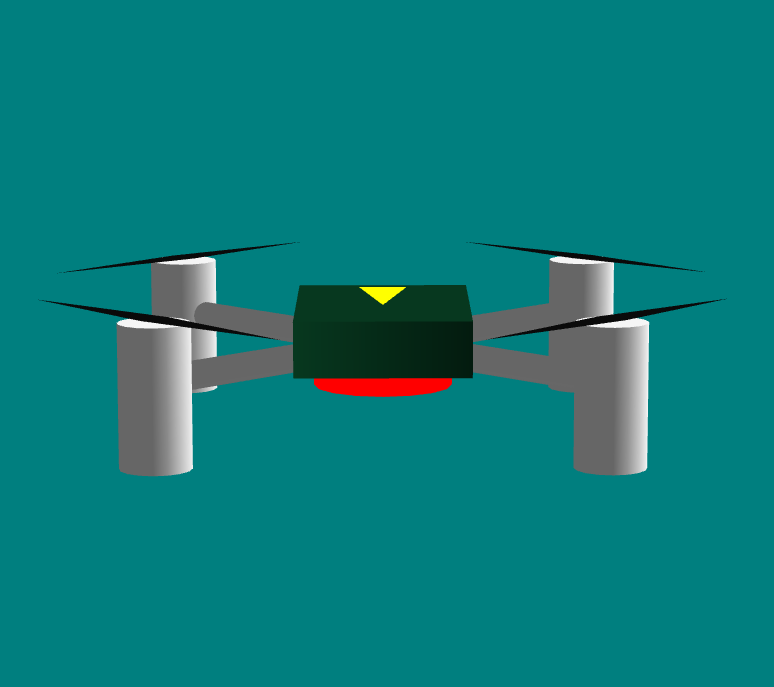}\label{fig:led_red}}
   \hfil
   \subfigure{\includegraphics[width=0.22\linewidth]{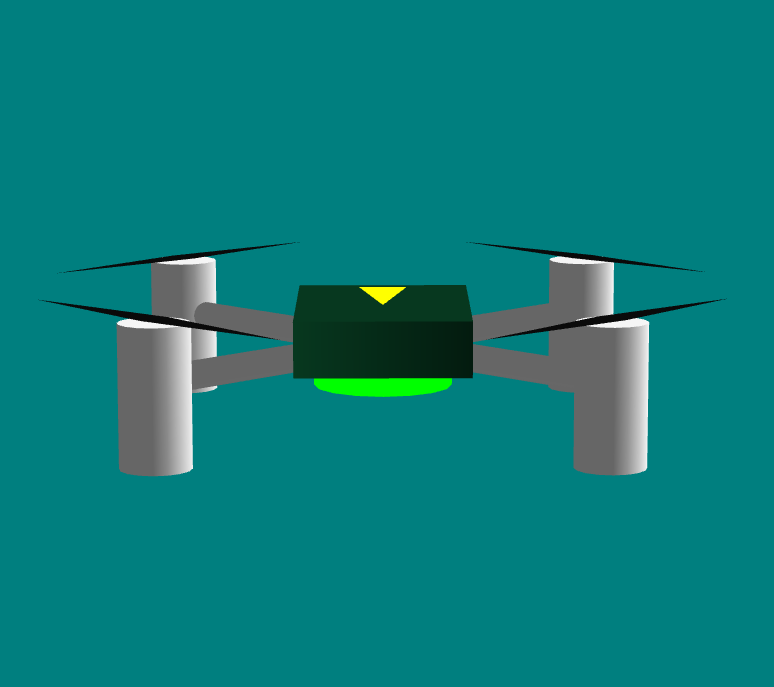}\label{fig:led_green}}
   \hfil
   \subfigure{\includegraphics[width=0.22\linewidth]{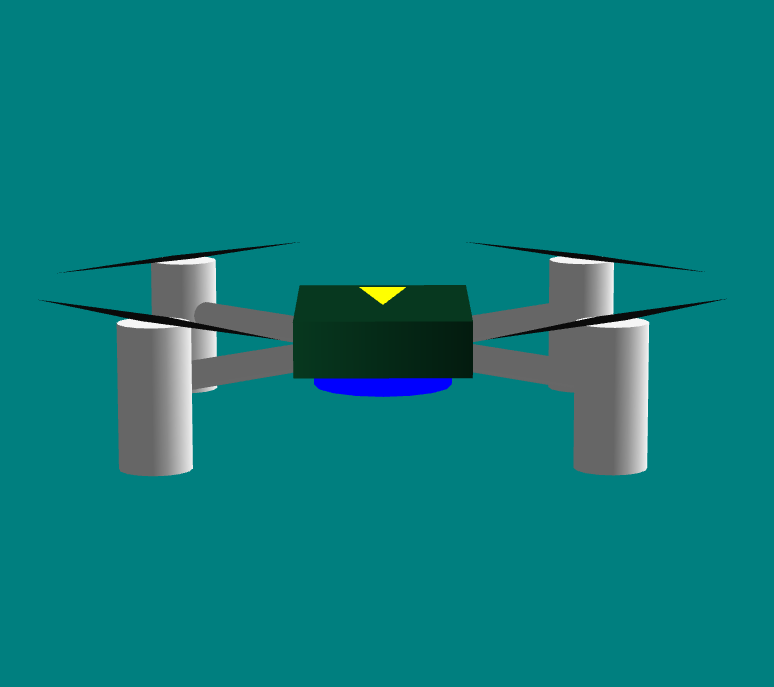}\label{fig:led_blue}}
   \\
   \subfigure{\includegraphics[width=0.22\linewidth]{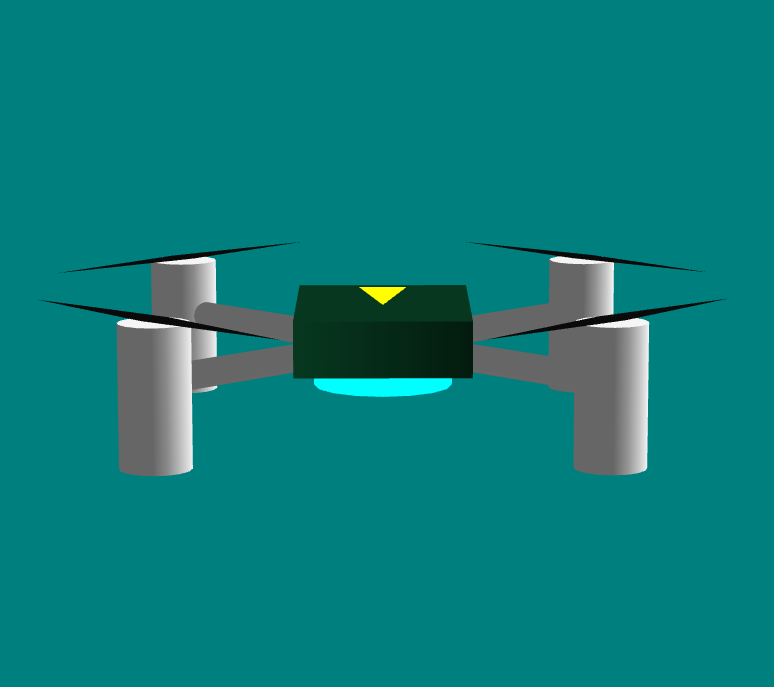}\label{fig:led_cyan}}
   \hfil
   \subfigure{\includegraphics[width=0.22\linewidth]{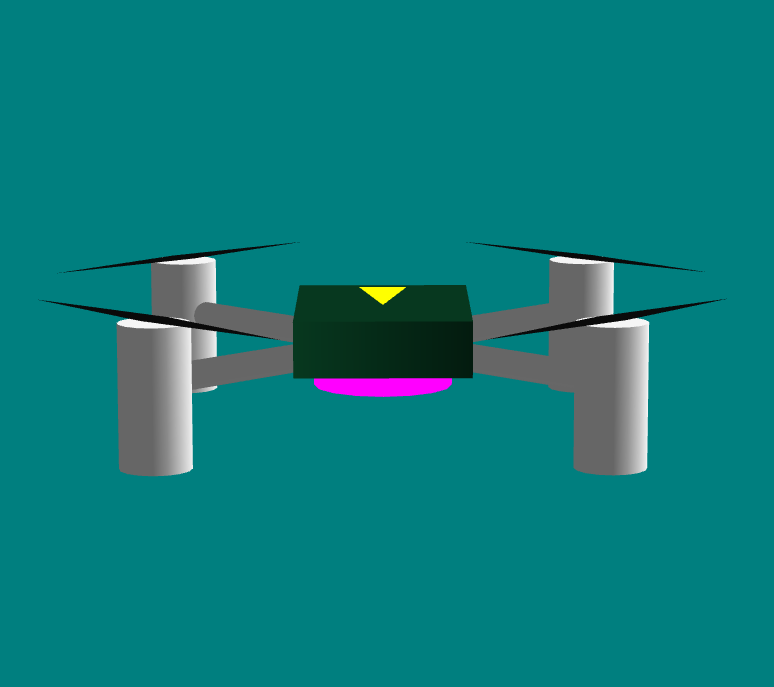}\label{fig:led_magenta}}
   \hfil
   \subfigure{\includegraphics[width=0.22\linewidth]{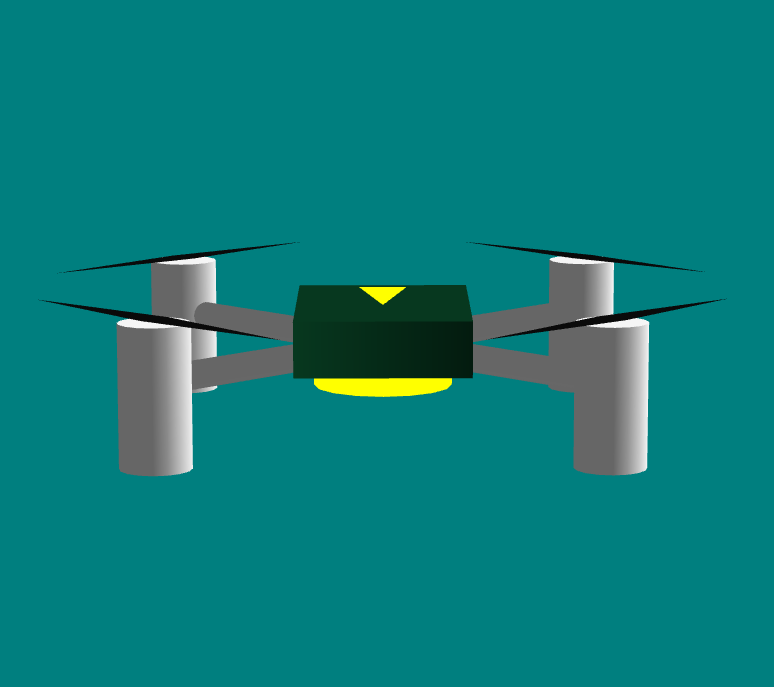}\label{fig:led_yellow}}
   \hfil
   \subfigure{\includegraphics[width=0.22\linewidth]{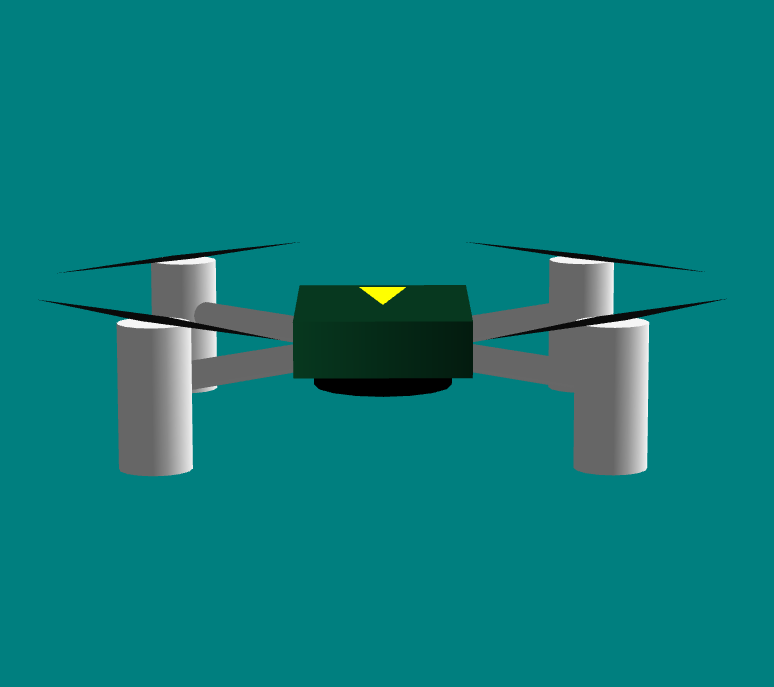}\label{fig:led_black}}
   \caption{Crazyflie 3D model showing the RGB LED included.}
   \label{fig:led}
\end{figure*}

The implemented drone body consists of the central board, four arms with their motors and propellers (static to save computing resources), and a bottom LED corresponding to the ring-deck.
The LED can be detected by the light sensors of other robots using the ARGoS' medium for LEDs.
This allows the iteration with other robots in the simulation without using the Range and Bearing sensors/actuators.
The LED can be lit in different colours by providing its RGB code (Figure~\ref{fig:led}).

\subsubsection{Onboard camera}

\begin{figure}[!t]
   \centering
   \includegraphics[width=0.95\linewidth]{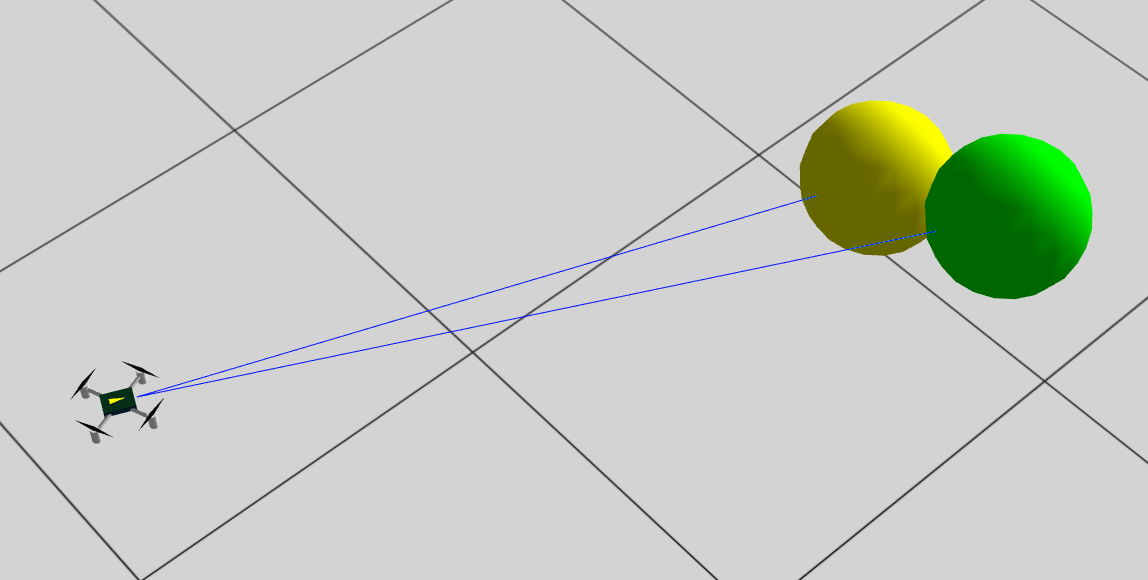}
   \caption{The implemented onboard camera detecting two light sources.}
   \label{fig:camera}
\end{figure}

We have implemented an optional onboard camera for the Crazyflie drone plug-in using the exiting ARGoS' perspective camera.
It simulates the available AI-deck for edge-computing capabilities.
It consists of a 320x320 greyscale camera (Himax HM01B0) for streaming images and implementing full autonomous flights.
Our implementation is able to detect ARGoS lights and LEDs (including the drone's onboard LED) and map them into a 320x320 coordinated plane to provide relative positions.
A graphical representation of the detection rays can be activated to help debugging simulations when using the ARGoS graphical visualisation (Figure~\ref{fig:camera}).

\subsection{Velocity and Position Controllers}

We have implemented a velocity PD controller to move the drone using a 3-axis speed vector allowing indicating relative or absolute speeds in metres per second.
Additionally, the existing position PD controller has been adapted to the Crazyflie's dynamics, being also possible to indicate a relative or absolute position in metres as the drone's destination.
Both controllers have been calibrated through simulation experiments and validated with real drones.

\subsubsection{Range and Bearing}

Like most of the ARGoS robots, the Crazyflie plug-in also supports the use of Range and Bearing communications, allowing the drones not only communicating with others robots but also detecting the direction and distance of the received communication signal.
Communication range can be restricted as well as the payload size.

\subsubsection{Battery}

Finally, a battery discharge model was calculated to simulate the Crazyflie's LiPo battery in both aspects, maximum autonomy and discharge curve.
It is also possible to define the initial charge of the battery.
We have collected data from real drones while hovering and flying at different speeds to characterise their battery's behaviour.

\section{Experiments and Results}
\label{sec:results}

Our experiments were thought to test the drone behaviour through simulations as well as validate trajectories and its battery discharge model using the real Crazyflie drones (Firmware 2023.07) and the ARGoS simulator (Version 3.0.0-beta59) running in a DELL XPS 15 9570 (12 Intel Core i7-8750H CPU @ 2.20 GHz and 16 GB of RAM).
We have set up 10 ticks per second as the simulation resolution (the default value).
For the real world experiments we have used the SwarmLab facility of the FSTM/DCS (University of Luxembourg) consisting of a 3x3x3-metre experimental area, two Lighthouse V2 base stations (positioning) and three Crazyflie drones.
The drones communicated with the laptop (via the Crazyradio PA 2.4 GHz USB dongle) where the controllers were implemented and the telemetry was recorded.

\subsection{Onboard camera}

\begin{figure}[!t]
   \centering
   \includegraphics[width=0.95\linewidth]{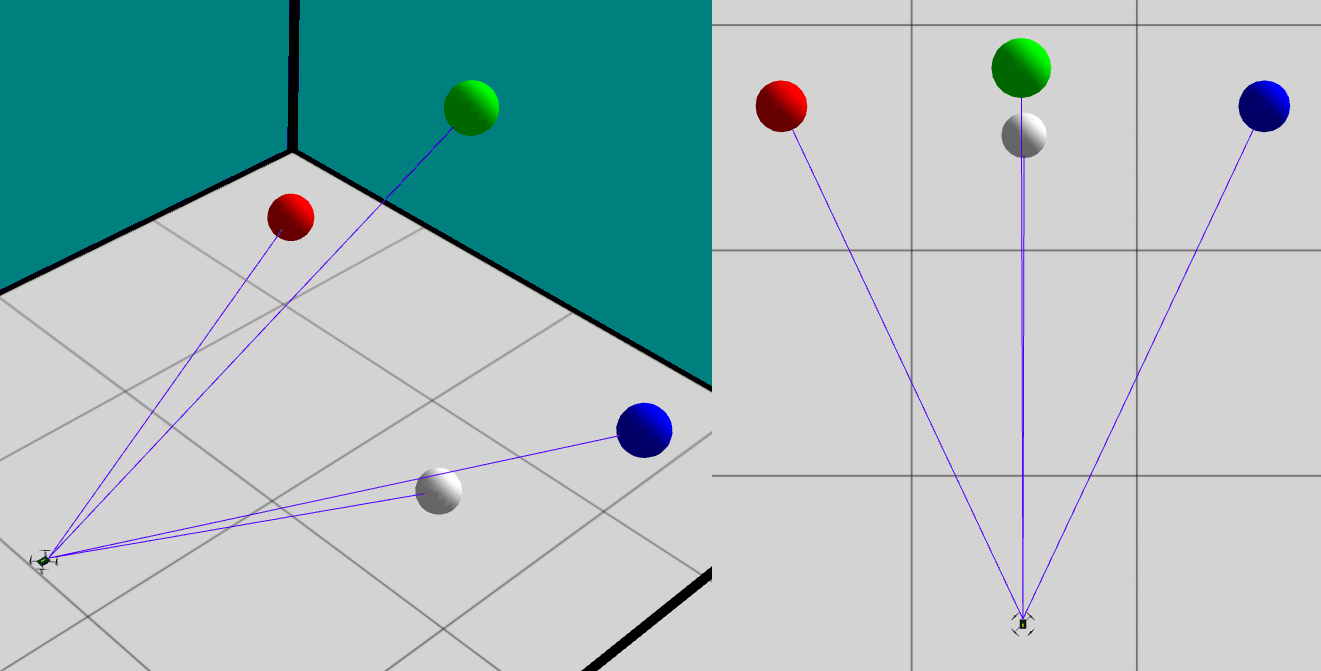}
   \caption{Calibration of the onboard camera. (50-degree aperture).}
   \label{fig:camera_calibration}
\end{figure}

We have calibrated the onboard camera following the AI-deck specifications.
Using the experimental setup shown in Figure~\ref{fig:camera_calibration} we have validated the implemented camera aperture.
The maximum distance between light sources was 1.8652 metres and they were placed 2 metres away from the Crazyflie's camera.
It allowed us to confirm that the maximum detection angle is indeed 50 degrees, matching the camera's aperture.
As the lights' coordinates are mapped in a 320x320 plane ([0 -- 319] range), we have obtained: (0,159) for red, (160,0) for green, (319,159) for blue, and (160,318) for white.
These values are according with the expected results.
Moving any of the light sources away from the detection area, would remove it from the detections reported by the onboard camera.

\subsection{Velocity Controller}

\begin{figure*}[!t]
   \centering
   \subfigure[0.25 m/s]{\includegraphics[width=0.24\linewidth]{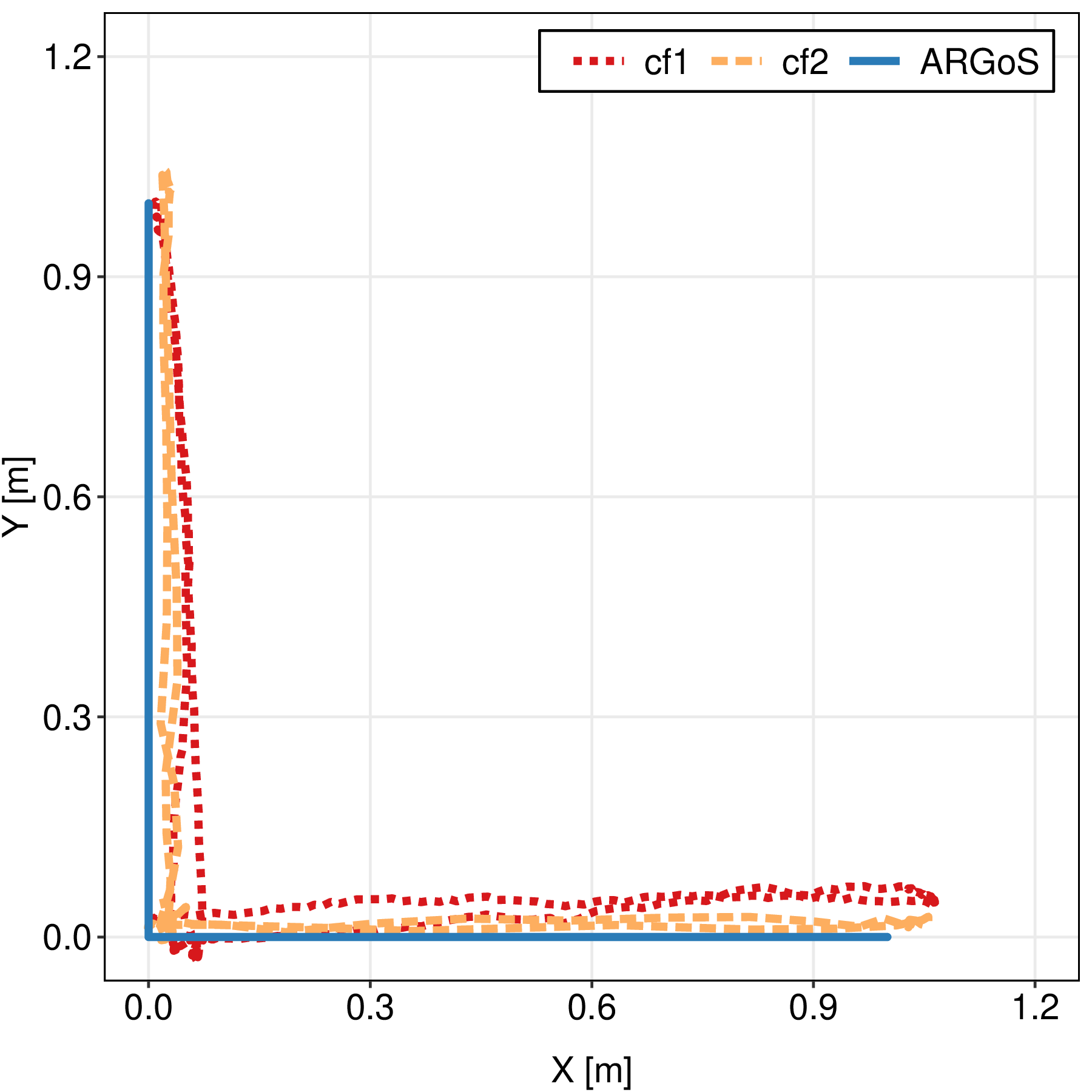}\label{fig:xy_025}}
   \subfigure[0.50 m/s]{\includegraphics[width=0.24\linewidth]{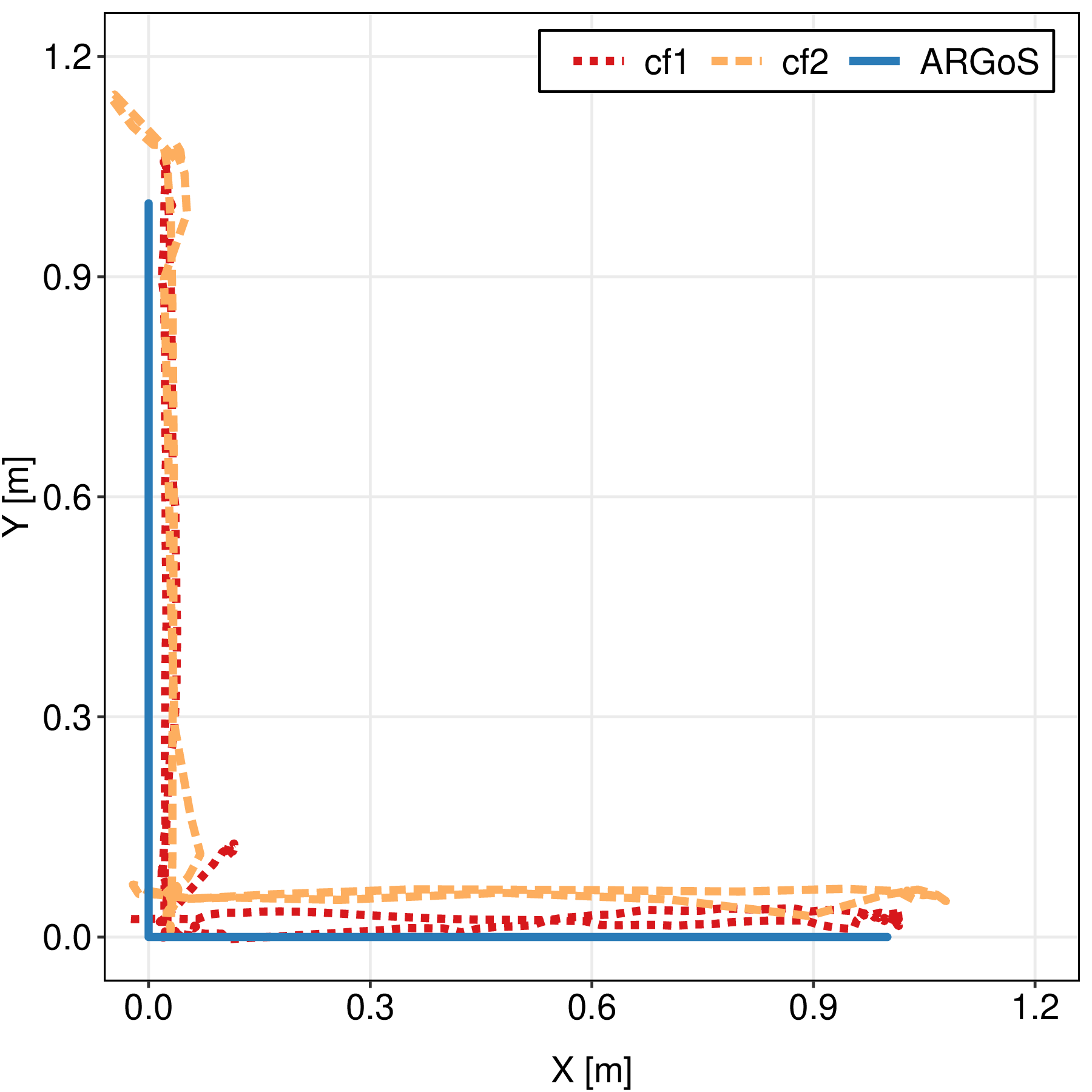}\label{fig:xy_050}}
   \subfigure[1.00 m/s]{\includegraphics[width=0.24\linewidth]{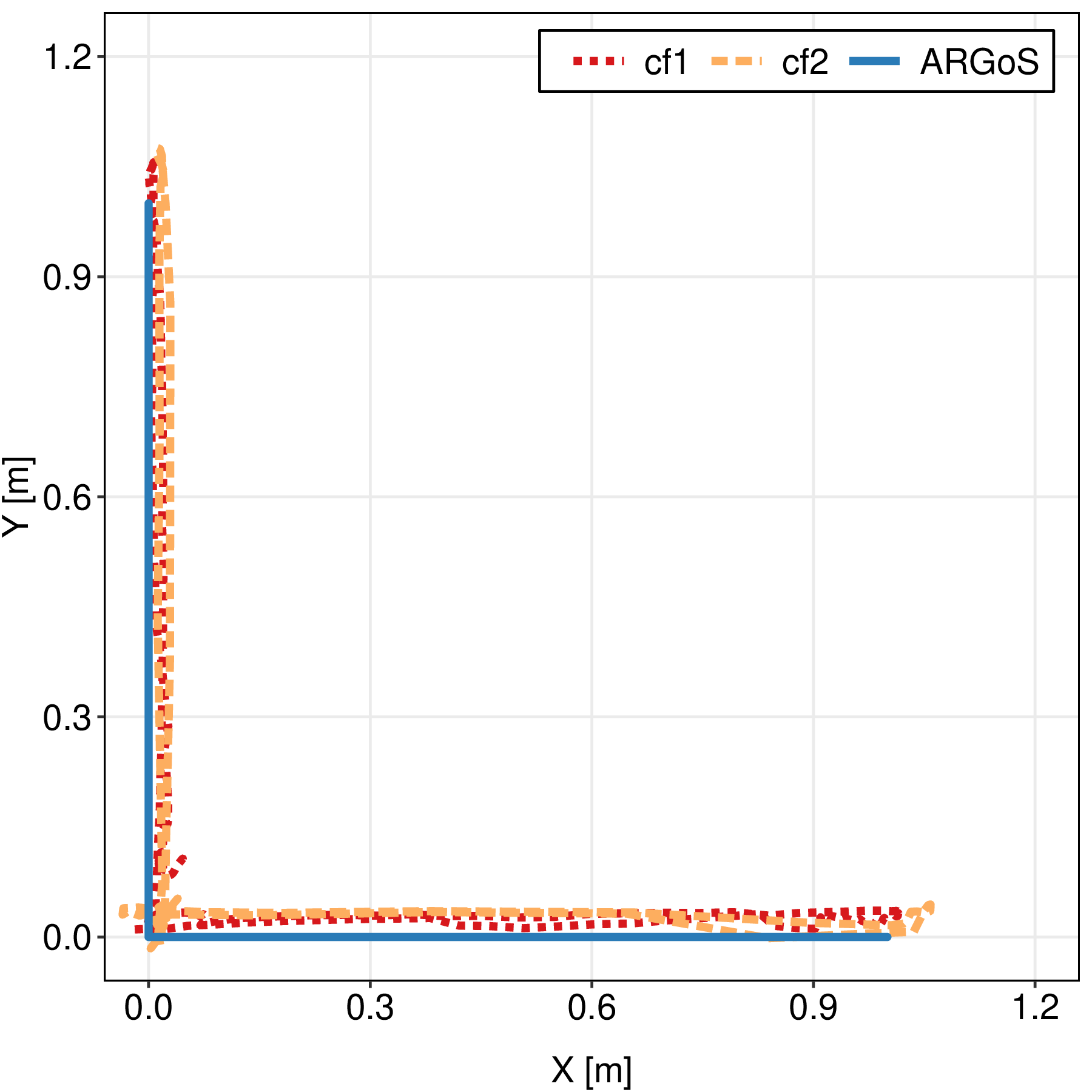}\label{fig:xy_100}}
   \caption{Comparison of 2D drone trajectories for two Crazyflie drones (cf1 and cf2) and the ARGoS plug-in.}
   \label{fig:xy}
\end{figure*}

The velocity PD controller was designed to simulate the dynamic drone behaviour observed from real world flights.
Firstly, we have set up a 2D flight plan following a straight trajectory along the x and y axes from the origin (0,0) to one metre away in both directions (0,1) and (1,0).
We have collected the trajectories from two Crazyflie drones (cf1 and cf2) flying a three different speeds: 0.25, 0.50, and 1.00 metre per second.
Figure~\ref{fig:xy} shows the plots of the drones' trajectories compared to our ARGoS' plug-in.
It can be seen that real world robots have an inherent low accuracy mainly due to the positioning system although air turbulences have been also observed.

\begin{figure}[!t]
   \centering
   \subfigure[0.25 m/s, x-y plane]{\includegraphics[width=0.48\linewidth]{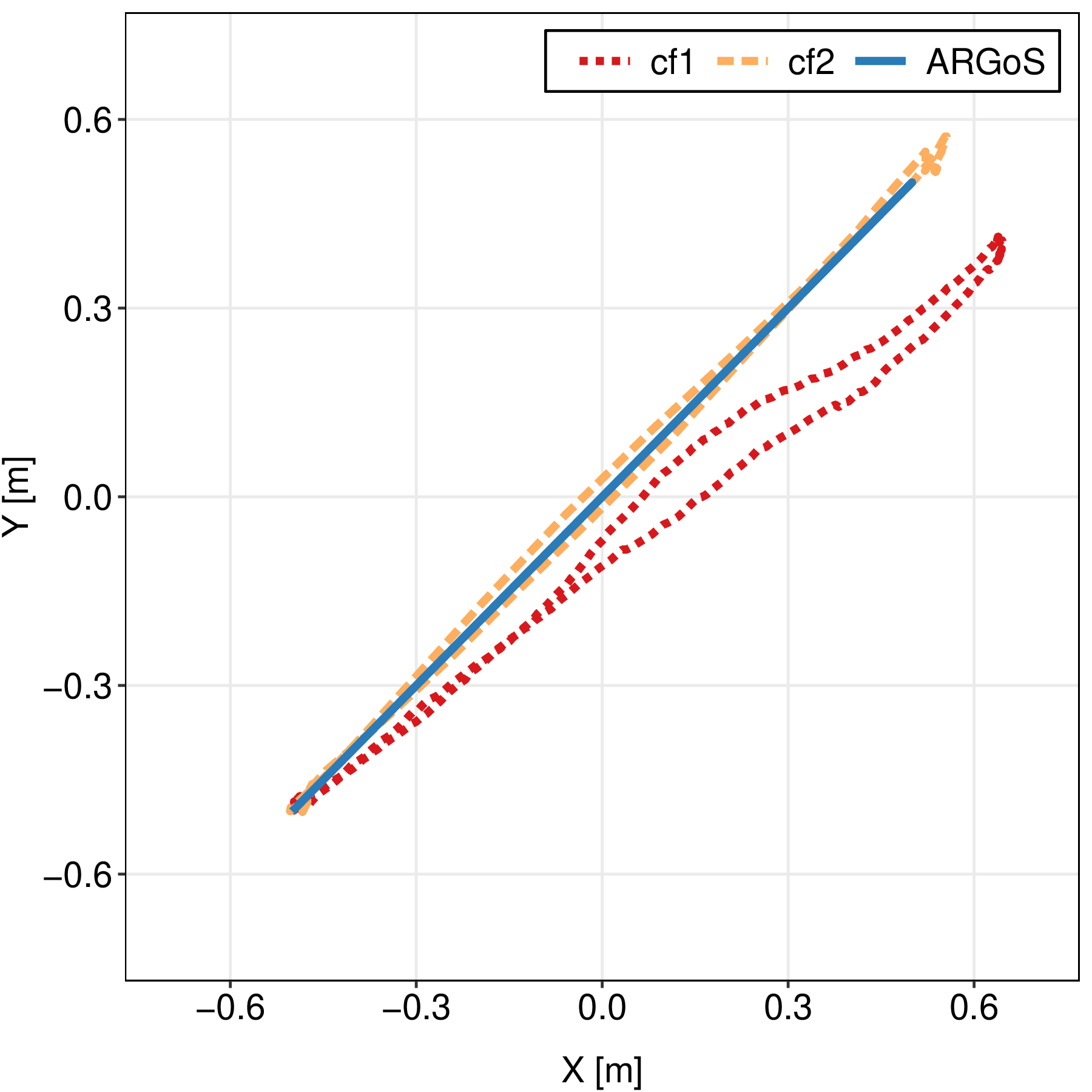}\label{fig:xyy_025-xy}}
   \hfil
   \subfigure[0.25 m/s, x-z plane]{\includegraphics[width=0.48\linewidth]{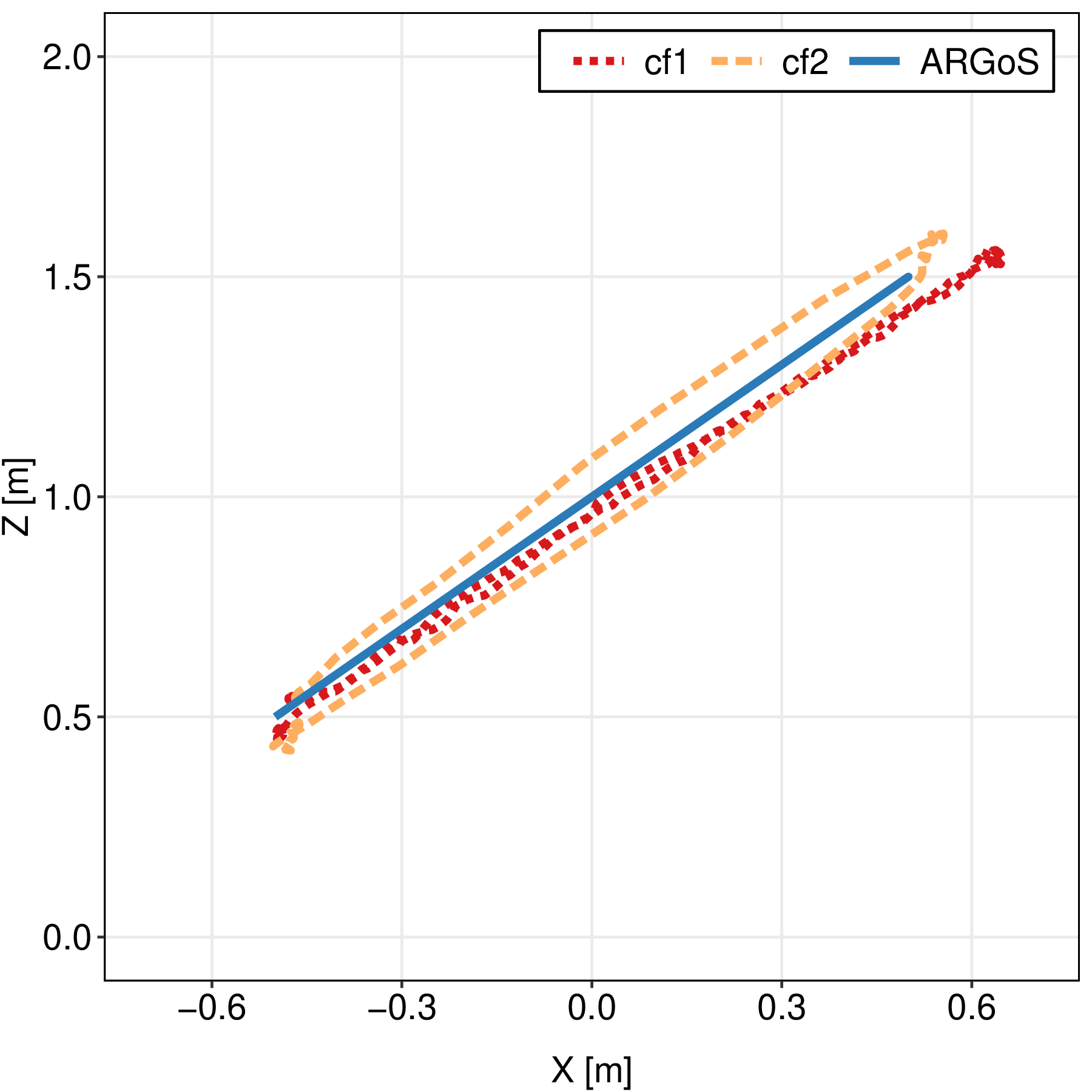}\label{fig:xyz_025-xz}}
   \\
   \subfigure[0.50 m/s, x-y plane]{\includegraphics[width=0.48\linewidth]{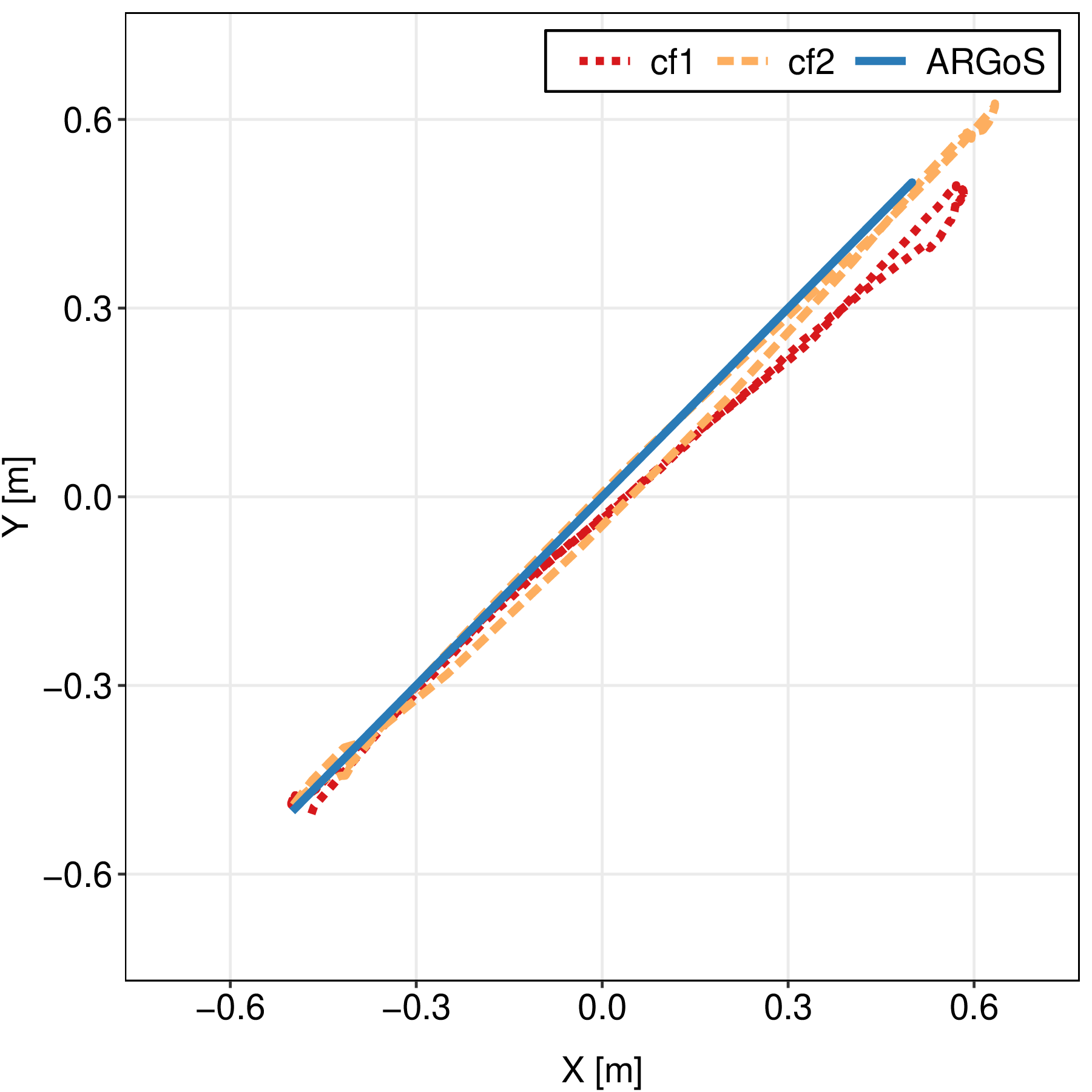}\label{fig:xyy_050-xy}}
   \hfil
   \subfigure[0.50 m/s, x-z plane]{\includegraphics[width=0.48\linewidth]{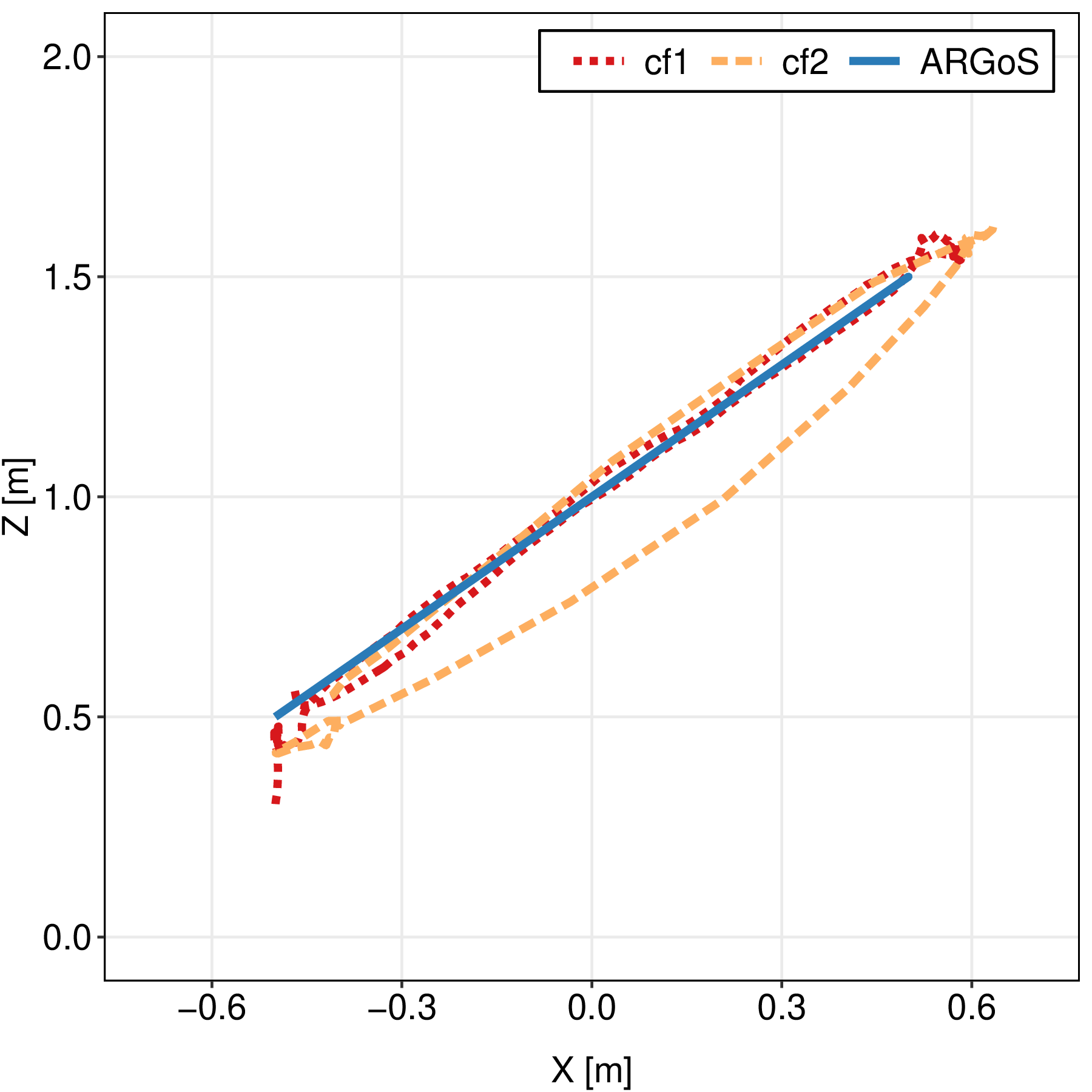}\label{fig:xyz_050-xz}}
   \\
   \subfigure[1.00 m/s, x-y plane]{\includegraphics[width=0.48\linewidth]{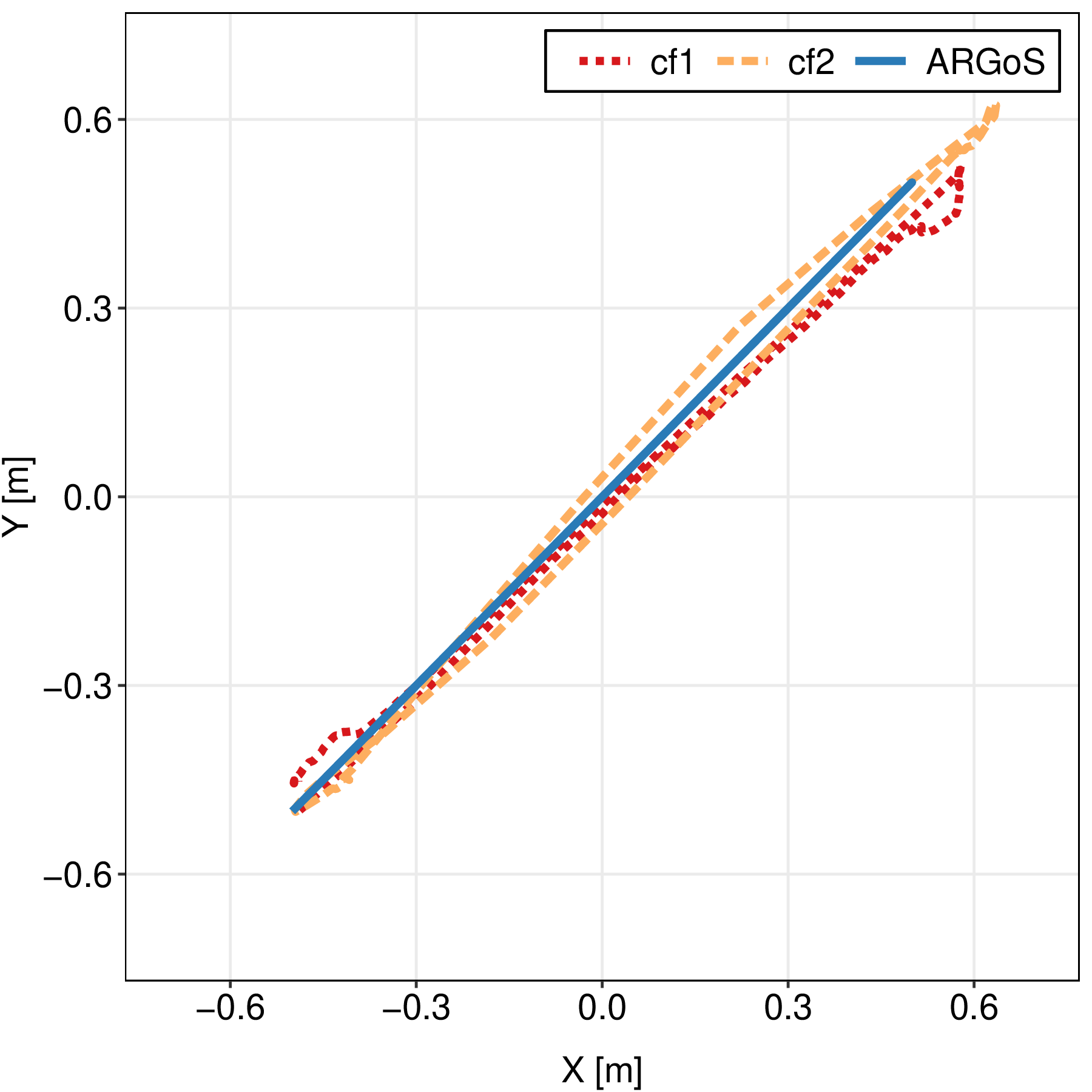}\label{fig:xyy_100-xy}}
   \hfil
   \subfigure[1.00 m/s, x-z plane]{\includegraphics[width=0.48\linewidth]{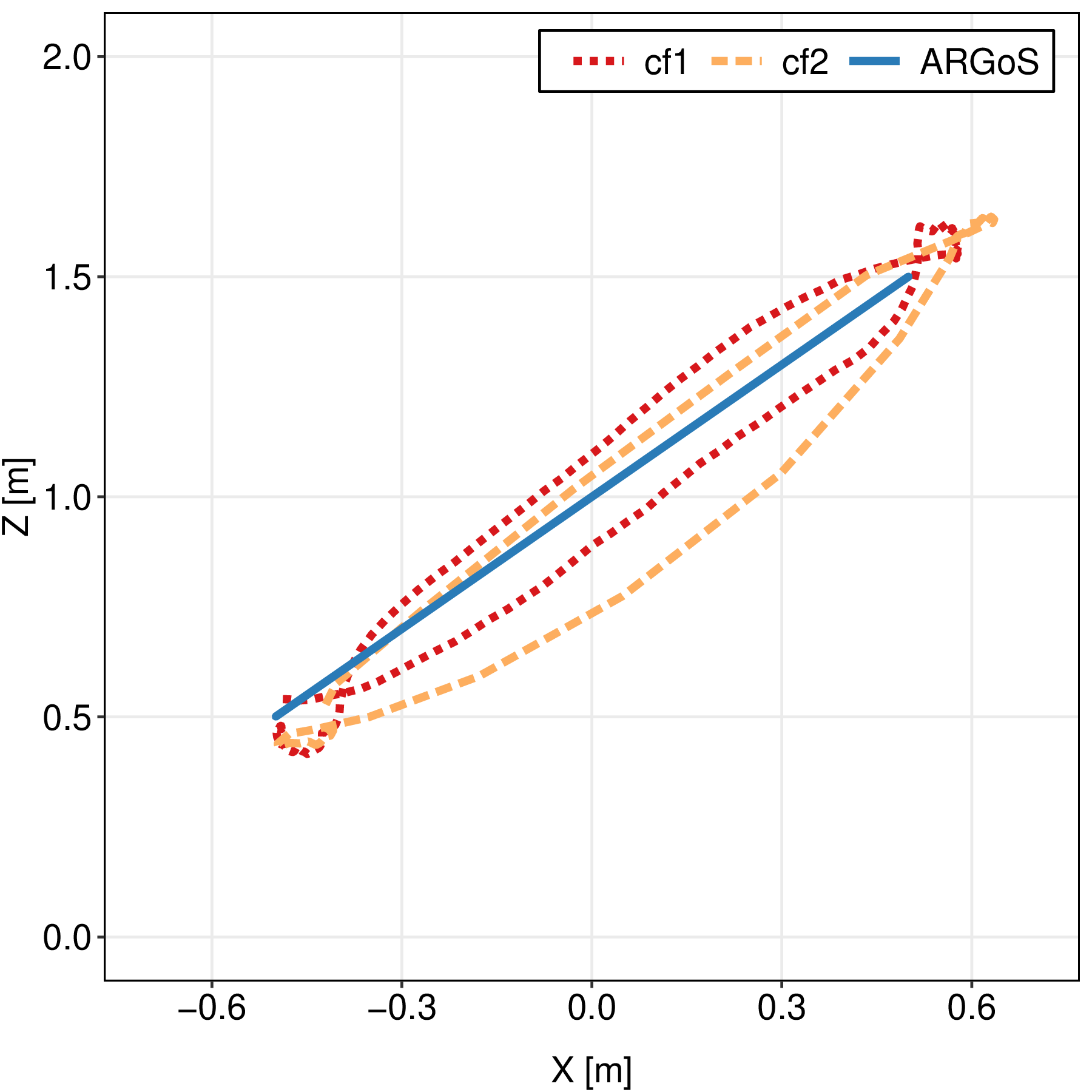}\label{fig:xyz_100-xz}}
   \caption{Comparison of 3D drone trajectories for two Crazyflie drones (cf1 and cf2) and the ARGoS plug-in.}
   \label{fig:xyz}
\end{figure}

Three-dimensional trajectories were also tested.
In this case the drones would fly form the (-0.5,-0.5,0.5) point to the (0.5,0.5,1.5) point, i.e. one metre along each axis.
The real drone trajectories as well as those collected from ARGoS for the three tested speeds, are shown in Figure~\ref{fig:xyz}.
It can be seen again that real drones lose accuracy when moving throughout the air.
Hence, we have conducted the experiment twice to increase the reliability of the collected data.

\begin{figure}[!t]
   \centering
   \subfigure[Altitude]{\includegraphics[width=0.48\linewidth]{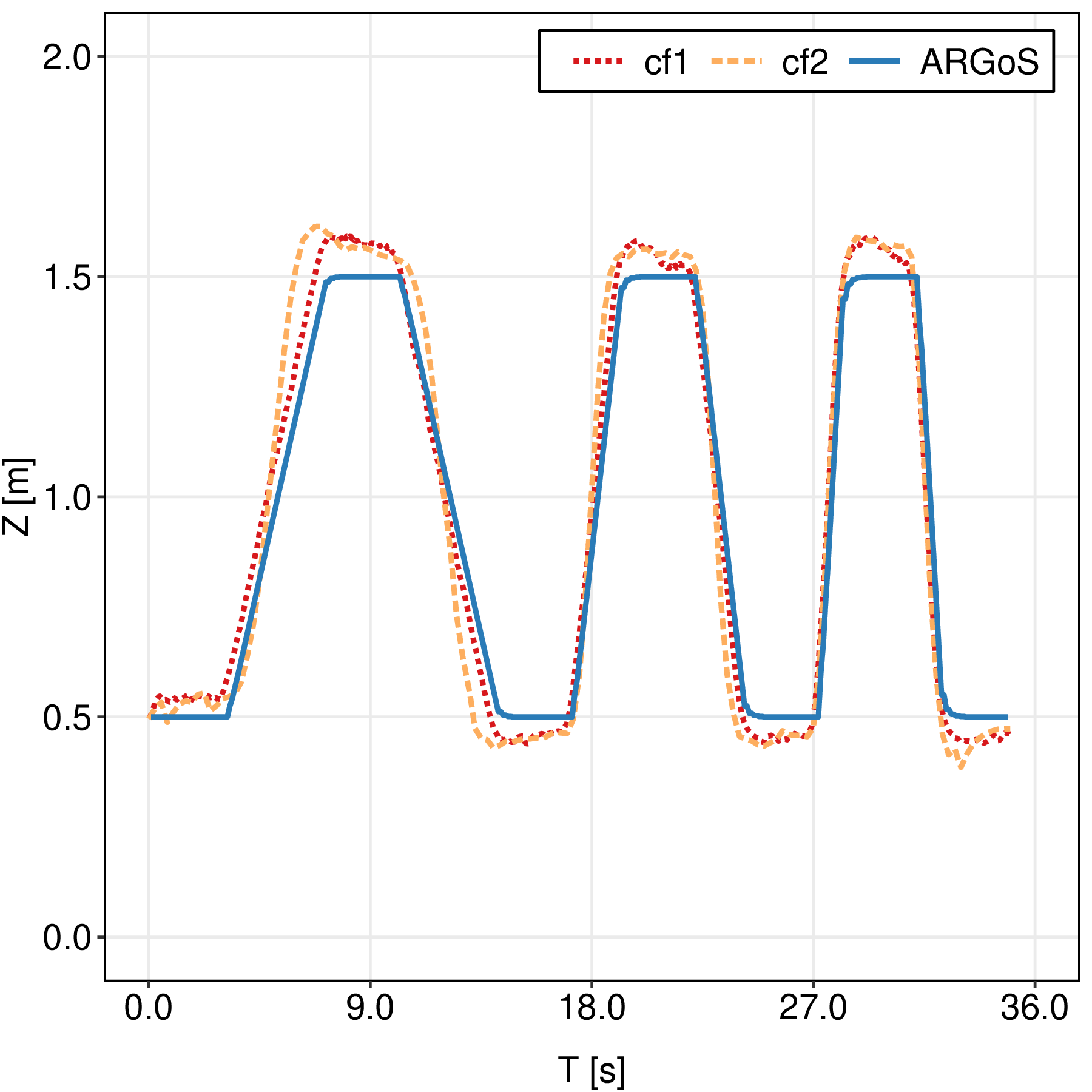}\label{fig:speed_z}}
   \hfil
   \subfigure[Yaw]{\includegraphics[width=0.48\linewidth]{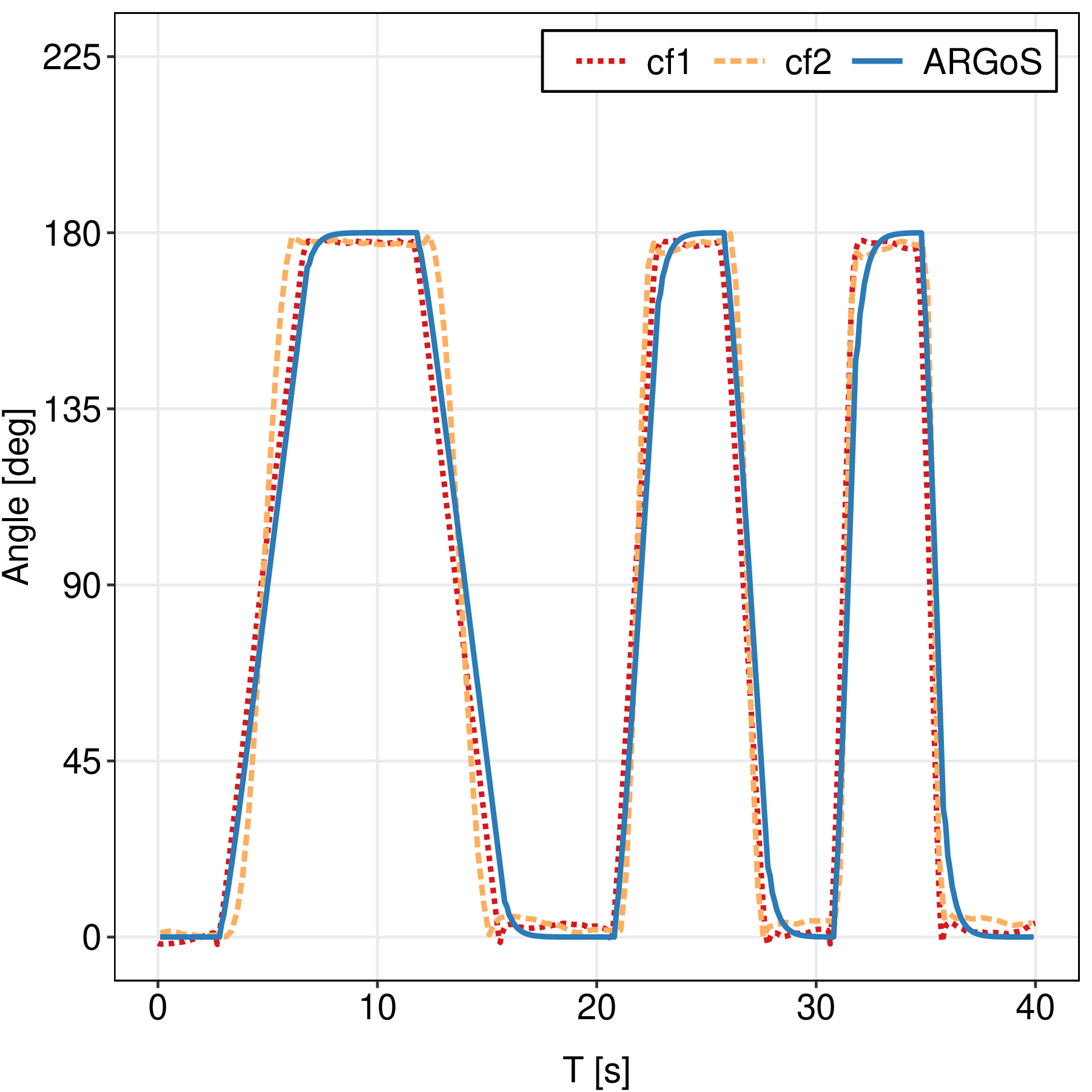}\label{fig:speed_w}}
   \caption{Comparison of altitude and yaw modifications for two Crazyflie drones (cf1 and cf2) and the ARGoS plug-in.}
   \label{fig:zw}
\end{figure}

Finally, we have tested different hovering altitudes and yaw modifications (z-axis rotations).
Figure~\ref{fig:speed_z} shows the evolution of the drones' altitude in time.
Again we have compared three different testing speeds using two Crazyflie drones and the simulator plug-in.
We can see that despite some bounces at the points in which the drones change their vertical speed, the real and simulated trajectories are close, denoting the plug-in accuracy.
Changes in the drone orientation (yaw) were also experimentally tested and emulated by our plug-in.
We can see in Figure~\ref{fig:speed_w} a comparison between two real drones and the ARGoS plug-in where changes of 180 degrees were tested.
Both behaviours, real and simulated, are almost identical as the real drones are more stable when their are just rotating.

All in all, we have observed that the velocity controller has accurately simulated the real Crazyflie behaviour in the ideal environment provided by a simulator.
The desired maximum speeds vs. the values measured from simulations are shown in Table~\ref{tab:speed_controller}.
We can see that the results are very accurate except for the maximal rotation speed tested (180 deg/s) which was not completely reached by the simulated drone (neither the real one).
We believe that the Crazyflie plug-in would have needed turning more that just 180 degrees to have time to reach that maximum rotational speed.

\begin{table*}[ht]
   \renewcommand{\arraystretch}{1.3}
   \caption{Accuracy of the speed controller.}
   \label{tab:speed_controller}
   \centering
   \begin{tabular}{ l | r r r | r r r | r r r | r r r}
     \hline 
     Experiment     & \multicolumn{3}{|c|}{Altitude [cm/s]}
                    & \multicolumn{3}{|c|}{Plane [cm/s]}
                    & \multicolumn{3}{|c|}{3D space [cm/s]}
                    & \multicolumn{3}{|c}{Yaw [deg/s]} \\
     \hline      
     Desired Max. Speed  & 2.5 & 5.0 & 10.0 & 2.5 & 5.0 & 10.0 & 4.33 & 8.66 & 17.32 & 45.0 & 90.0 & 180.0 \\
     Measured Max. Speed & 2.5 & 5.0 & 10.0 & 2.5 & 5.0 & 10.0 & 4.33 & 8.66 & 17.32 & 45.0 & 89.6 & 170.6 \\
     \hline      
     Error          & 0.0 & 0.0 & 0.0 & 0.0 & 0.0 & 0.0 & 0.0 & 0.0 & 0.0 & 0.0 & 0.4 & 9.4 \\
     \hline      
   \end{tabular}
\end{table*}

\subsection{Position Controller}

\begin{figure}[!t]
   \centering
   \subfigure[Trajectory]{\includegraphics[width=0.48\linewidth]{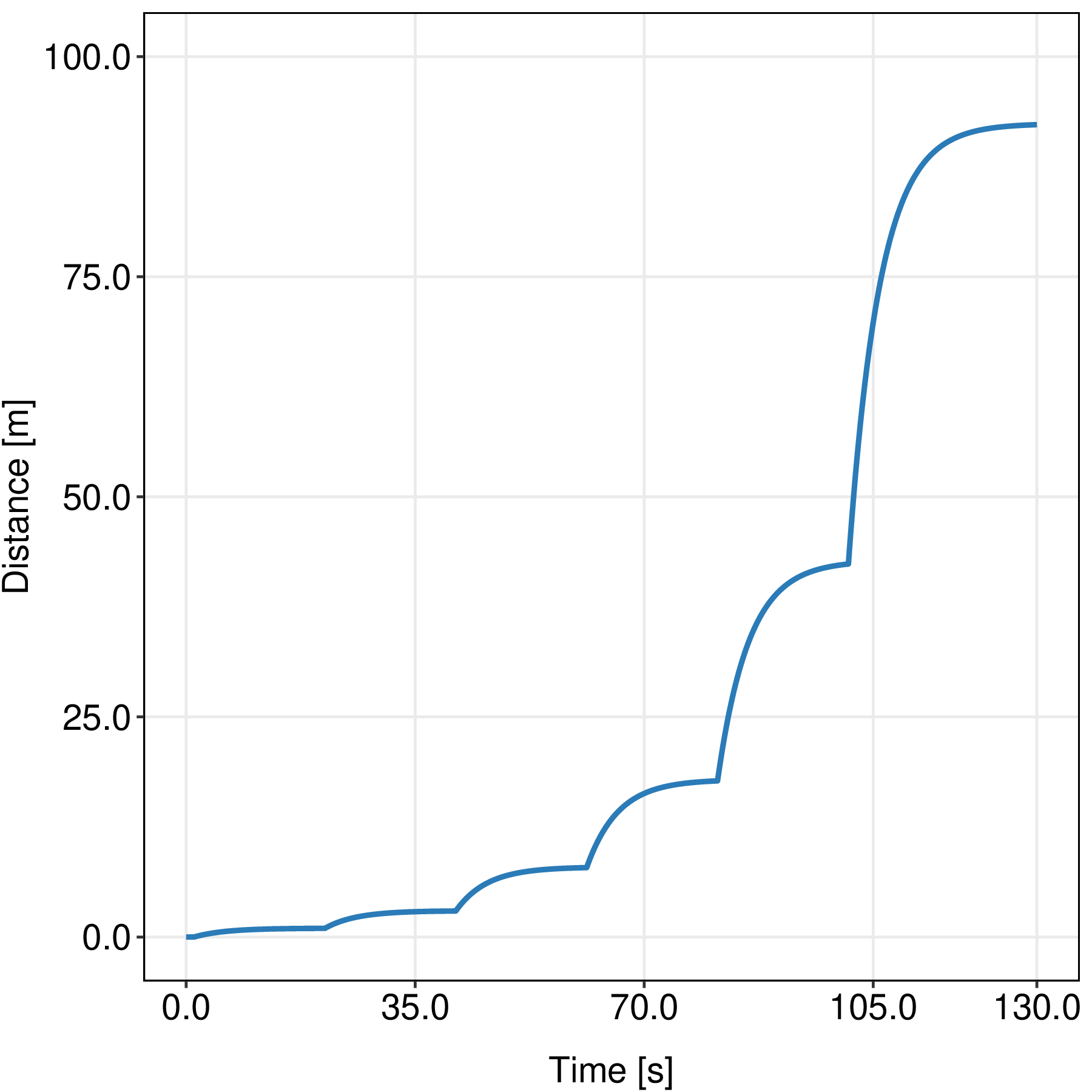}\label{fig:pos_x_pos}}
   \hfil
   \subfigure[Linear Speed]{\includegraphics[width=0.48\linewidth]{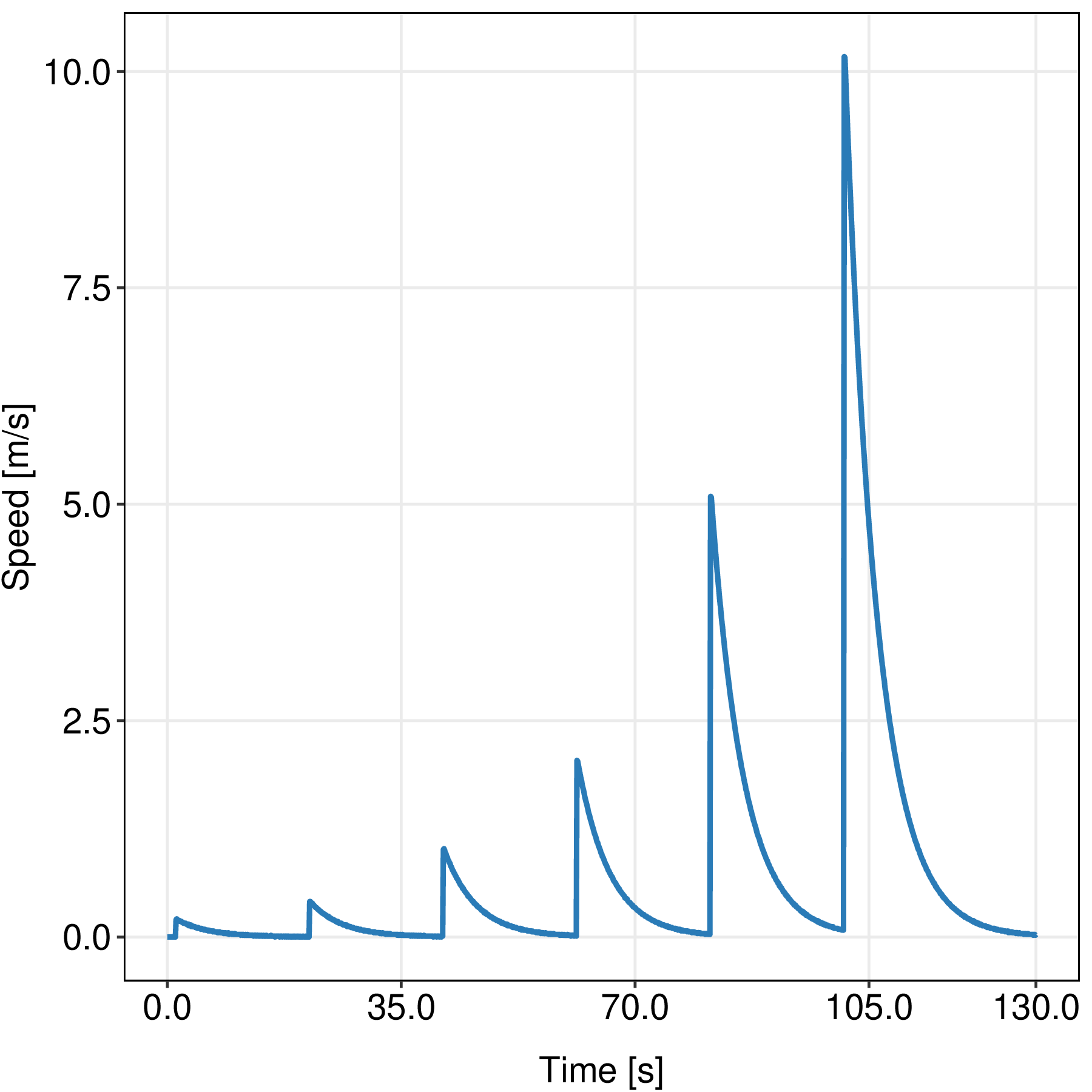}\label{fig:pos_x_spd}}
   \\
   \subfigure[Yaw]{\includegraphics[width=0.48\linewidth]{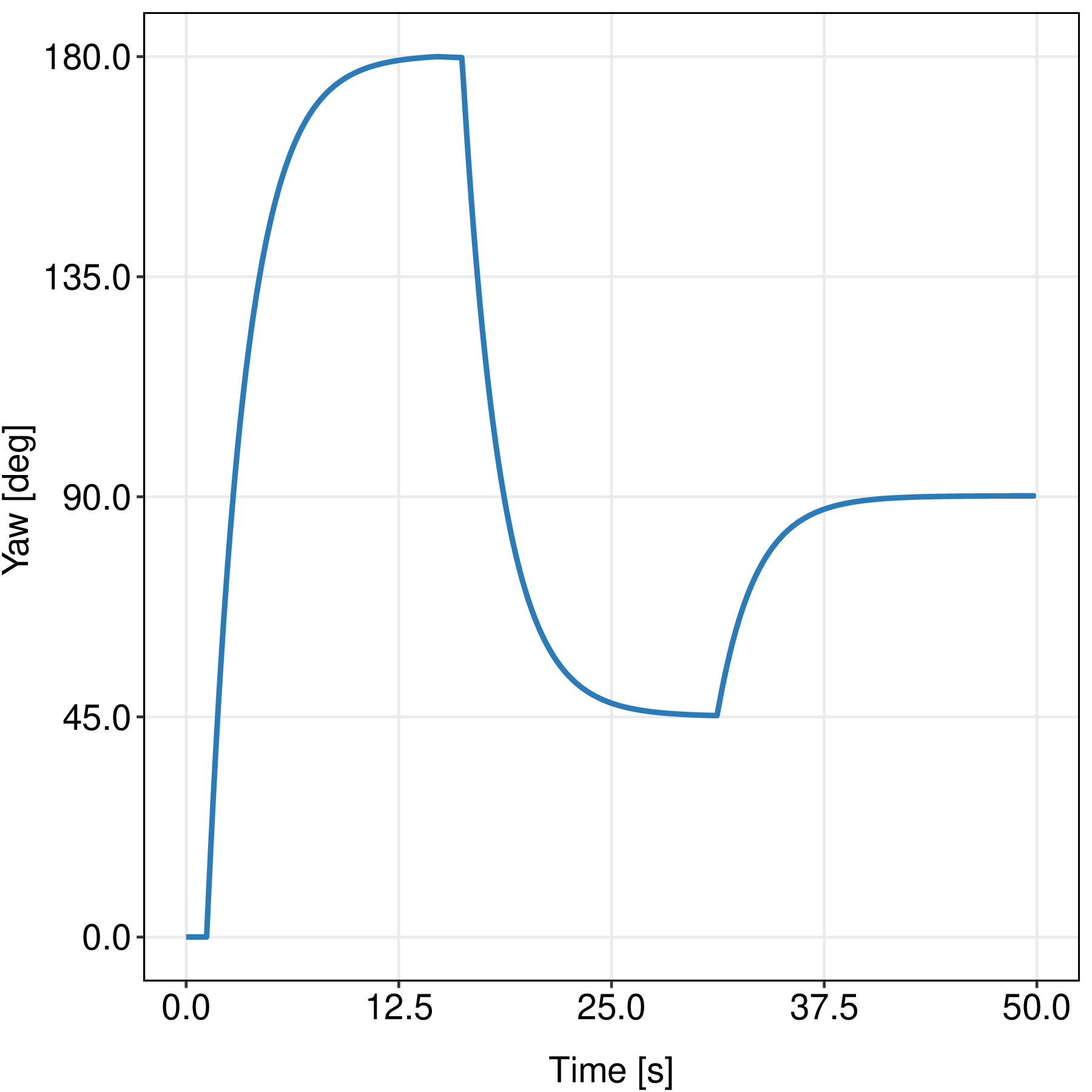}\label{fig:pos_w_pos}}
   \hfil
   \subfigure[Radial Speed]{\includegraphics[width=0.48\linewidth]{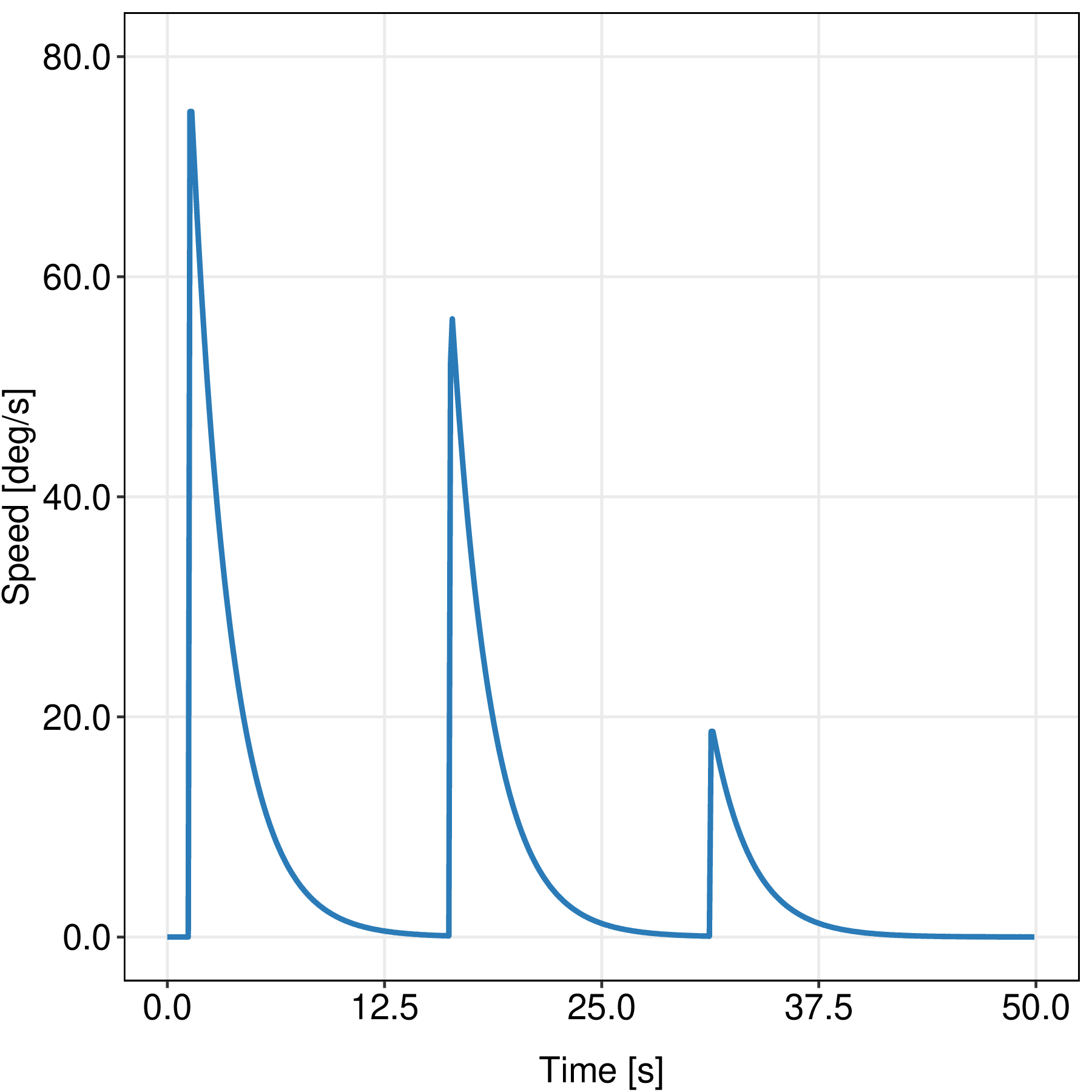}\label{fig:pos_w_spd}}
   \caption{Trajectories, rotation angle and speeds for the ARGoS plug using the position controller.}
   \label{fig:position}
\end{figure}

We have modified the ARGoS' existing position controller to adapt its behaviour to the Crazyflie drone.
We have set up two experiments to assess its position and speed accuracy when moving in a linear trajectory and also when rotating in a static position.
Figure~\ref{fig:pos_x_pos} shows the trajectory obtained when the simulated drone moved 1, 2, 5, 10, 25, and 50 metres.
It can be seen in Figure~\ref{fig:pos_x_spd} that the maximum speed is limited to 10 metres per second.
Moreover, three changes in the drone's yaw have been tested.
The simulated Crazyflie has been rotated 180, 135, and 45 degrees to test its accuracy (Figure~\ref{fig:pos_w_pos}) and maximum rotational speed (Figure~\ref{fig:pos_w_pos}).
Alongside the rotational accuracy it can be seen that the maximum speed was set to 90 degrees per second (never reached during the experiment).

Table~\ref{tab:position_controller} shows the results of the position controller experiments.
It can be seen that the desired travelling distance was never completely reached.
We believe that the observed errors come from the fact that the drone is beginning the next move without having enough time to reach the intermediate destinations.
It is supported by the accuracy of the short distance displacements and also by the last one, where after a 50-metre trajectory, the drone only was 7 centimetres away of the desired distance.
Similar results were observed for the rotations tested where the intermediate value presented the higher error, although all of them were lower than 1 degree.

\begin{table*}[ht]
   \renewcommand{\arraystretch}{1.3}
   \caption{Accuracy of the position controller.}
   \label{tab:position_controller}
   \centering
   \begin{tabular}{ l | r r r r r r | r r r }
     \hline 
     Experiment     & \multicolumn{6}{|c|}{Trajectory [m]}
                    & \multicolumn{3}{|c}{Yaw [deg]} \\
     \hline      
     Desired Distance  & 1.00 & 2.00 & 5.00 & 10.00 & 25.00 & 50.00 & 180.00 & -135.00 & 45.00 \\
     Measured Distance & 0.98 & 1.97 & 4.92 &  9.85 & 24.62 & 49.93 & 179.84 & -134.54 & 44.90 \\
     \hline      
     Error             & 0.02 & 0.03 & 0.08 &  0.15 &  0.38 &  0.07 &   0.16 &    0.46 &  0.10 \\
     \hline      
   \end{tabular}
\end{table*}

\subsubsection{Battery}
\label{sec:results_battery}

\begin{figure}[ht]
   \centering
   \subfigure[Data collected from three drones and the polynomial fitted]{\includegraphics[width=0.48\linewidth]{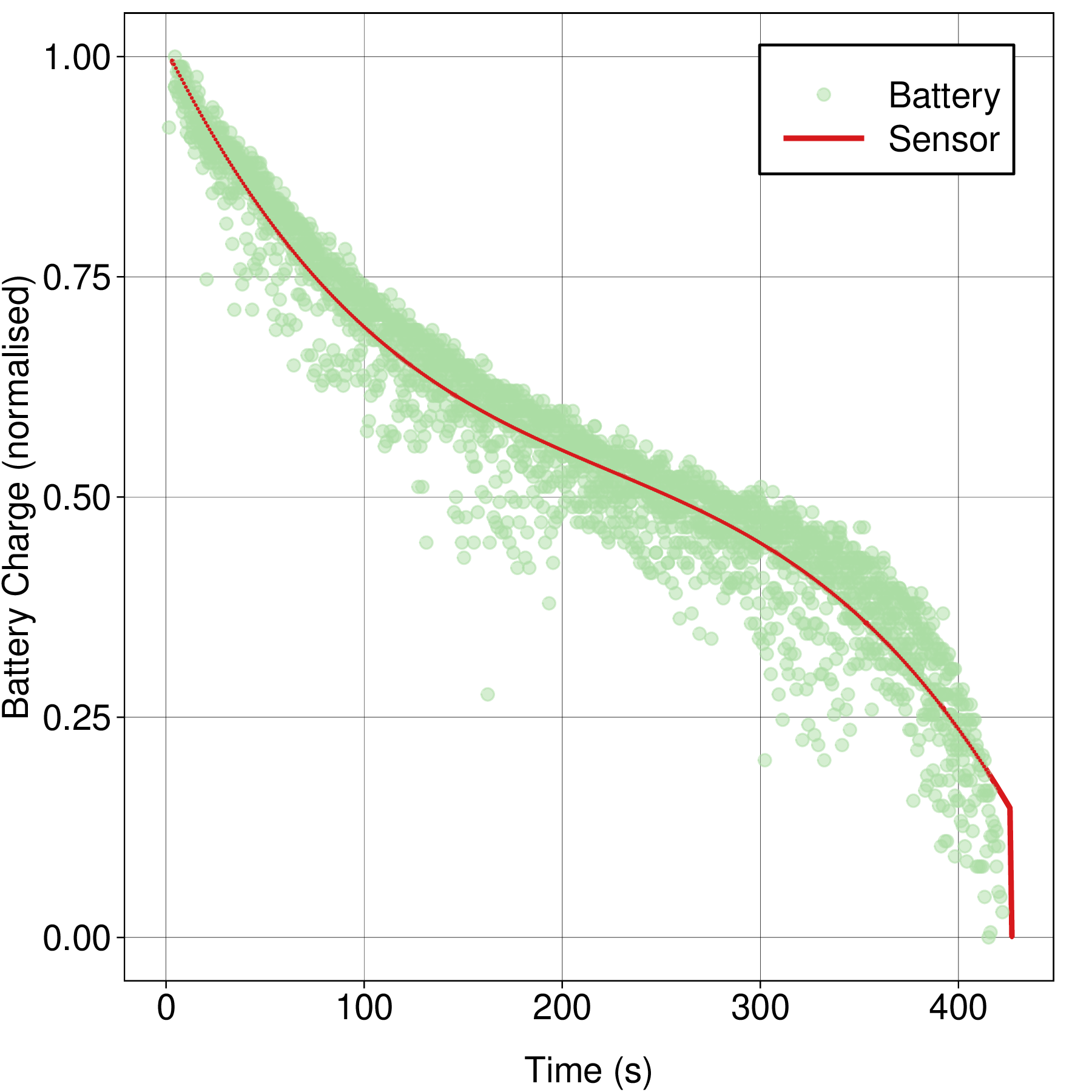}\label{fig:battery_data}}
   \hfil
   \subfigure[Polynomial fitted (Sensor) and the data collected from ARGoS]{\includegraphics[width=0.48\linewidth]{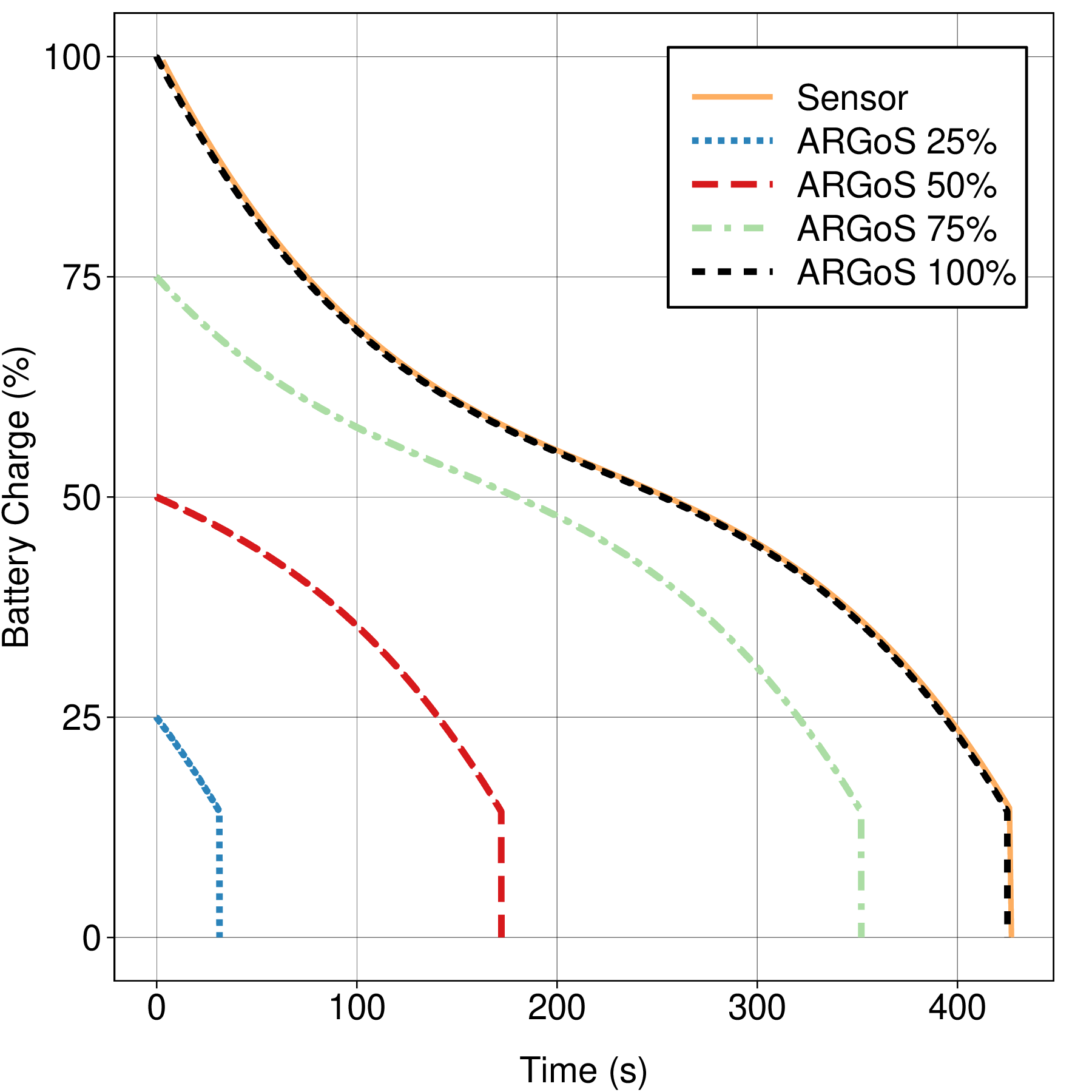}\label{fig:battery_argos}}
   \caption{Data collected from the battery experiment, third-degree polynomial fitted to the points, and data collected from the implemented ARGoS plug-in.}
   \label{fig:battery_experiment}
\end{figure}

The battery model in ARGoS relies on an associated discharge model.
We have evaluated three Crazyflie drones moving at different speeds as the motors are the main source draining the battery.
Firstly, we have collected the individual data points measured from the battery charge and fitted a third-degree polynomial to those points, while keeping the observed maximum flying time, i.e. 427.21 seconds (Figure~\ref{fig:battery_data}).
Note that this sharp reduction of the battery charge corresponds to the value in which the remaining energy is not enough to keep the drone flying.
Secondly, we have implemented this battery discharge model in the ARGoS plug-in and tested it through simulations to assess if the polynomial discharge was correctly represented (Figure~\ref{fig:battery_argos}).
Additionally, different initial battery charges, i.e. 25\%, 50\%, 75\%, and 100\%, were tested to verify the correct behaviour of the discharge model.

The implemented discharge algorithm is presented in Algorithm~\ref{alg:discharge}.
Firstly, it calculates the corresponding $t$ for the discharge curve based on the $CurrentCharge$ (the positive real root of the cubic polynomial).
Secondly, if the maximum $t$ ($Tmax$) was reached, the $NextCharge$ value is set to 0 (battery depleted).
Otherwise, the provided $\Delta_t$ is used to calculate the next $t$ value.
Finally, the $NextCharge$ is calculated using the cubic polynomial $P$.

\begin{algorithm}
   \caption{Battery Discharge Model.}
   \label{alg:discharge}
   \begin{algorithmic}
      {\small
      \Function{Cubic}{$\Delta_t, CurrentCharge$}
         \State $t \gets FindRoot(P, CurrentCharge)$ \Comment{Finds t}
         \If {$t \geq Tmax$}
            \State $NextCharge \gets 0$ \Comment{Maximum t was reached}
         \Else
            \State $t \gets t + \Delta_t$
            \State $NextCharge \gets P(t)$ \Comment{Next charge value for $t$}
         \EndIf        
         \State \textbf{return} $NextCharge$ \Comment{Next battery charge value}
      \EndFunction
      }
   \end{algorithmic}
\end{algorithm}

We have compared the battery charge values obtained with the collected values from the drones in order to assess its accuracy.
We have used the Mean Square Error (MSE) as a metric, calculated as shown in Equation~\ref{eq:mse}, where $n$ is the number of data points, $Y_i$ are the observed values, and $\widehat{Y_i}$ are the estimated values.
We have obtained $MSE = 0.002114$, a very low error value that denotes the high accuracy of the fitted model.

\begin{equation}
   MSE = \frac{1}{n}\sum_{i=1}^{n}(Y_i - \widehat{Y_i})^2\label{eq:mse}
\end{equation}

\section{Conclusion}
\label{sec:conclusion}

In this article we have presented the Crazyflie drone plug-in for the ARGoS simulator.
We have described the new graphic model for the robot and implemented its sensors, controllers and actuators, and tested the results obtained to assess their accuracy and fidelity.
We have modelled the new drone body and added the optional expansion decks such as the LED ring and the onboard camera (AI-deck).
We have adapted the Spiri position PD controller to this drone and also implemented a new speed PD controller to allow experimenting by setting up a constant flight speed.
Finally, we have implemented a new battery discharge model to restrict the Crazyflie's flying time by using a realistic battery's discharge curve.

We have conducted experiments for measuring and validating the plug-in's trajectories compared to real drones.
We have calibrated the battery using real discharge data and measured the MSE of our implementation to confirm its accuracy. 
We have observed precise results for the speed PD controller and some deviations on the final positions achieved by the position PD controller.
We believe that part of them are occurring because we have not given the drone enough time to reach de final desired position at then end of some intermediate trajectories.

As a matter of future work we would like to further test our plug-in and evaluate the possibility of implementing a full PID (proportional-integral-derivative) controller to minimise the final error observed in some of our experiments.

\section{Acknowledgements}

This work is supported by the Luxembourg National Research Fund (FNR) -- ADARS Project, ref. C20/IS/14762457.
The experiments presented in this paper were carried out using the the SwarmLab facility of the FSTM/DCS.

\bibliographystyle{elsarticle-num} 
\bibliography{crazy}

\end{document}